\documentclass[letterpaper, 10 pt, conference]{ieeeconf}  % Comment this line out
 \usepackage[]{times}                                                         % if you need a4paper
\IEEEoverridecommandlockouts                              % This command is only
\usepackage{amsmath} % assumes amsmath package installed
\usepackage{amssymb}  % assumes amsmath package installed
\usepackage{clrscode}
\usepackage{graphicx,subfigure}

\title{\LARGE \bf
Emotional Metaheuristics For in-situ Foraging Using Sensor
Constrained Robot Swarms }

\author{Esh Vckay and Debasish Ghose% <-this % stops a space
\thanks{Esh Vckay is a graduate student in the department of Electrical and Computer Engineering,
        The University of Texas at Austin, Austin, TX 78712, USA
        {\tt\small eshvk@utexas.edu}}%
\thanks{Debasish Ghose is a professor at the Guidance, Control and Decision Systems Laboratory in the Department of Aerospace Engineering, Indian Institute of Science, Bangalore 560012, India
        {\tt\small dghose@aero.iisc.ernet.edu}}%
}

\begin{document}

\maketitle
\thispagestyle{empty}
\pagestyle{empty}

%%%%%%%%%%%%%%%%%%%%%%%%%%%%%%%%%%%%%%%%%%%%%%%%%%%%%%%%%%%%%%%%%%%%%%%%%%%%%%%%
\begin{abstract}

We present a new social animal inspired emotional swarm intelligence
technique. This technique is used to solve a variant of the popular
collective robots problem called foraging. We show with a simulation
study how simple interaction rules based on sensations like hunger
and loneliness can lead to globally coherent emergent behavior which
allows sensor constrained robots to solve the given problem.

\end{abstract}

%%%%%%%%%%%%%%%%%%%%%%%%%%%%%%%%%%%%%%%%%%%%%%%%%%%%%%%%%%%%%%%%%%%%%%%%%%%%%%%%
\section{INTRODUCTION}

  Foraging \cite {1balch1994communication} is a collective robotics problem that derives biological inspiration
  from the behavior of ants \cite{2h�lldobler1990ants}. Ants engaged in foraging, scout for prey, recruit nest mates
  when prey has been located, and work together as a group to bring back food to the nest. Foraging belongs to a
  class of problems known as coverage problems \cite{3choset2001coverage}. Typically, the objective of such
  problems is to ensure that certain areas of interest within a search space are explored by one or many robots.
  Applications of such problems include lawn mowing \cite{4huang1986region}, harvesting \cite{5ollis1996first},
  and mine removal \cite{6land1998coverage}.

     Solutions to such problems can involve one or multiple robots. Multi-robot approaches \cite{7zheng2005multi}
      \cite{8rekleitis2001multi} are particularly attractive because of the significant saving in time. However,
      designing provably complete solutions which require a methodological search of the environment come
      with drawbacks in the form of expensive sensors and computational resource requirements
      \cite{9vanderheide13117terrain}. Moreover, shortcomings during practical implementation
      like sensor errors makes it difficult to guarantee completeness. In such situations, solutions
      that are based on random search techniques \cite{10gage1993randomized} and do not require expensive sensors become just as effective.

     Biologically inspired techniques such as swarm intelligence utilize this philosophy of inexpensive design to solve coverage problems
     in general \cite{11rutishauser2009collaborative} and more particularly many variants of the foraging problem \cite{12labella2006division}.
     Swarm intelligence \cite{13Bonabeau} uses metaphors adopted from the behavior of social insects to provide solutions to complex engineering problems. Typically, these
     solutions are found to be adaptive, robust and scalable \cite{14sahin2005swarm}. In this paper, We adopt the same design goals of swarm intelligence especially the idea of self-organization, i.e. we try to ensure that a coherent global pattern can emerge from the local interactions of a system's constituent units. In this paper, we propose a metaheuristic that derives inspiration from human emotions to solve a variant of the foraging problem which we will call an in-situ foraging problem. The goal here is to utilize a swarm of robots to remove certain
     objects of interest which we will refer to as prey distributed in a two-dimensional search space which will be referred to as the field. In particular, we take an evolutionary perspective \cite{15cosmides2000epa} of
     emotions as mechanisms that have evolved to ensure survival of the species. Specifically, we use hunger and loneliness as a basis to design rules of interaction for the swarm. The paper is organized as follows: In the next section, we first present the biological foundations that our metaheuristic is founded upon. We continue by describing the metaheuristic in detail and a broader description of the different behaviors exhibited by the robots. Next, we present a study of the performance of our algorithm viz a viz two control algorithms. We then conclude by discussing results generated using a sensor based simulation model \cite{14sahin2005swarm} and make certain comments on future directions for this work.

 \section{BIOLOGICAL FOUNDATIONS}
 Evolutionary psychologists \cite{15cosmides2000epa} believe that certain actions and behaviors  observed in humans do not necessarily require complicated cognitive decision making mechanisms.  They are to be considered as specific programs that have evolved to deal with certain recurrent problems. Furthermore, they hypothesize the existence of a 'super program' - emotion that is designed by natural selection to select, coordinate and recalibrate one subset of many mutually exclusive behaviors (sub programs) so that the net output is to optimally deal with the specific problem and maximize the individual's chances of survival. We adopt this philosophy of emotions in the design of local decision rules which decide the individual's emotional state that maximizes the success of the system in completing the given task.

   Hunger is an emotional drive that ensures that a living being eats the appropriate amount of food for survival.  We normally associate it with two states: \emph{Hungry} and \emph{Satiated}. A state change occurs depending on the time spent till now without food and the amount of capacity left. We adopt this idea to design a rule which decides whether a robot should invite other robots to the location of a prey or not. The change of state from satiated to hungry determines whether the robot will act altruistically or more selfishly to minimize cooperation.

    An evolutionary viewpoint of loneliness is that of a sense of isolation that ensures that an individual seeks out other members of the same species (conspecifics) whether to reproduce or to get protection from predators \cite{16cacioppo-loneliness}. Inspiration is taken from this to allow robots to decide whether they will stop or re-issue invitations depending on whether they find company they find at the location of the prey.

    A combination of both biologically plausible rules control the level of cooperation which takes place during grazing. This has the advantage of ensuring that the number of robots that graze at a particular area is not externally specified by a global controller or some sort of sophisticated coordination mechanism. This form of local self-organized behavior becomes even more relevant in critical applications such as clearing of nuclear spills where electronic sensors don't work properly due to the effect of radiation \cite{17messenger1992effects}.
  \section{METAHEURISTICS}
 \subsection{Behaviors}
    Each robot engages in two or more subtasks that depending on the heuristic used, the foraging problem can be decomposed into:
    \begin{itemize}
    \item{\textbf{Random search:} The robot looks for prey randomly exploring the field within the given time interval. Once prey is found, the robot settles down to graze (or remove the prey). If the robot is recruited or invited to graze at a particular location, it switches to directed search behavior. }
    \item{\textbf{Directed Search:} The robot searches for prey in a specific direction along which a broadcast signal is being sent. It continually moves towards the broadcast signal unless it locates a closer signal or locates any prey. }
       \item{\textbf{Grazing:} Once the robot locates a prey which it is able to do only when it is right at the location of the prey, it stops and settles to remove the prey at a fixed rate, $R_{G}$ till its container is completely full. Once this happens, It shuts down operation.}
           \item{\textbf{Inviting:} In order to communicate that it has found something of interest, A robot broadcasts pheromones which are dispersed within a certain radius. This enables other robots which can detect the pheromones to engage in a directed search which leads them towards the source of the pheromone broadcast, i.e. the inviting robot. The decision to invite is dependent on the metaheuristic involved.}

    \begin{figure}[b]
    \vspace{-.5cm}
      \centering
      \includegraphics[scale=0.35,angle=-90]{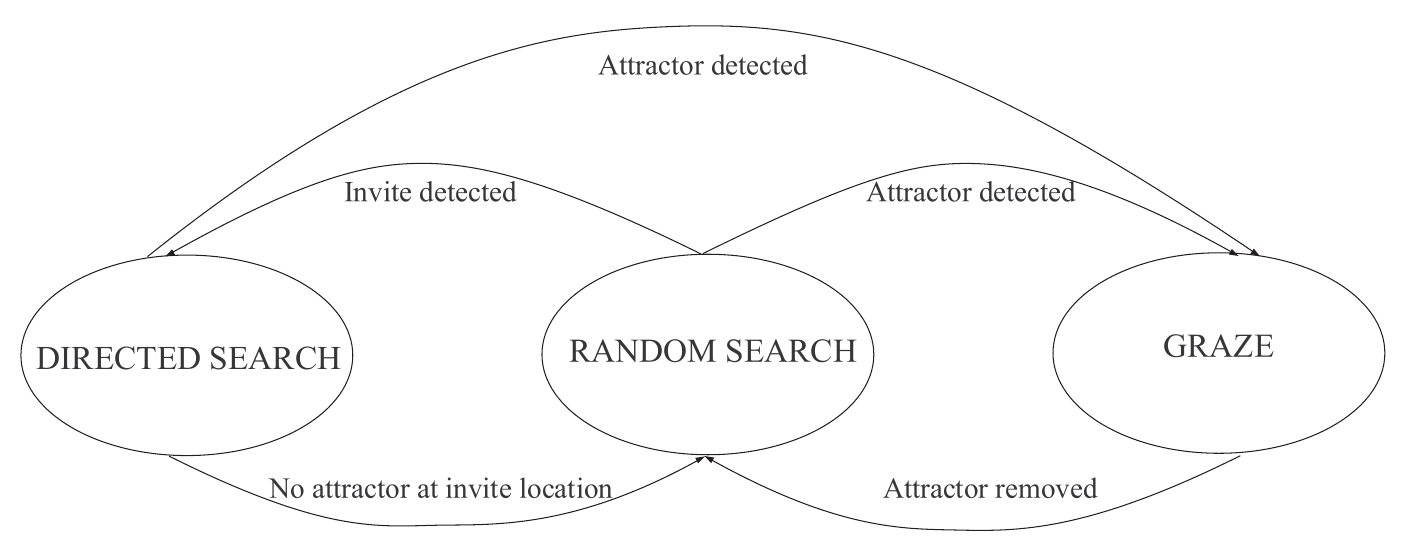}
      \caption{Overview of robot behaviors.}
      \label{fsm}
   \end{figure}

    \end{itemize}
    \subsection{The hunger-loneliness metaheuristic}
    Every robot's interaction in a time step with the rest of the swarm is determined by its current emotional state. This is in turn dependent on the value of Hunger ($\mathbf{H}$) and Loneliness (L parameters which are both bounded within a given interval $[1,100]$. The two parameters are updated on the basis of the following rules:
    \begin{itemize}
    \item{\textbf{Hunger update rule:} Increase or decrease $\mathbf{H}$ within the given bounds depending on whether a robot has grazed prey in the current time step or not. }
    \item{\textbf{Loneliness update rule:} Increase or decrease $\mathbf{L}$ within the given bounds depending on whether a grazing robot is within a certain proximity of other grazing robots. Keep increasing $\mathbf{L}$ when the robot is not grazing in the current time step. }
    \end{itemize}
    Every robot is in one of four emotional states, depending on the value of $\mathbf{H}$ and $\mathbf{L}$. The robot sends an invite only when its hunger is satiated and its loneliness is high ($\mathbf{H}_{L}\mathbf{L}_{H}$). Hunger is satiated ($\mathbf{H}_{L}$) when $1\leq\mathbf{H}\leq 50$ and is high $(\mathbf{H}_{H})$ when $50\le\mathbf{H}\leq100$. Similarly, the robot has low loneliness $\mathbf{L}_{L}$, when $1\leq\mathbf{L}\leq50$ and high loneliness ($\mathbf{L}_{H}$), when $50\le\mathbf{L}\leq100$. In a realistic scenario involving robots that have fixed battery power, energy can be conserved by selectively performing invites to a particular location.

The advantage of this is that in a  realistic scenario involving robots that have limited capacity to graze prey
    \subsection{The random search metaheuristic}
    This is the first of two control metaheuristics which we use to compare the performance of our algorithm against. Each robot independently searches for prey and grazes when it finds them. However, no robot transmits invitation signals and there is no explicit cooperation with other robots.
    \subsection{The random search-immediate invite metaheuristic}
    This control metaheuristic places a high emphasis on cooperation to the extent that each robot starts sending out invite signals once it immediately starts grazing at the spot where prey is found. Robots that respond to invite engage in a directed search towards the point where the prey is present.

\section{EMPIRICAL STUDY}
\subsection{Experimental setup}
   A simple particle model is used to simulate the robots. We don't consider any kinematic or dynamic constraints on the robots in this study. The robots placed in the field are free to move out of it while conducting a random search. Constraints on the robots include a fixed invite broadcast range and that even if a robot detects prey, it will not be able to determine the size of the prey. Each robot also has a fixed container size capacity of $100$ with a grazing rate of $R_{G} = 1$. During the random search, the robot engages in a two-dimensional random walk \cite{18weissteinrandomwalk} where its trajectory in $\mathbf{N}$ time steps is composed of $\mathbf{N}$ two-dimension vectors with random orientations and fixed magnitude $\mathfrak{l}$. For such a random walk, the root-mean-square distance traveled after  $\mathbf{N}$ steps of length $\mathbf{l}$ is given by $d_{\emph{rms}} = \mathbf{l}\sqrt{\mathbf{N}}$. $d_{\emph{rms}}$ is particularly significant in the design of the environment such that the spatial distribution of prey is such that the robots can ideally cover the entire area within the allocated time of $\mathbf{N}$ steps. Also, each prey covers the same spatial area with their densities determining the number of robots required to remove the prey.

   \subsection{Heuristic performance measure}
     Similar to other analyses of foraging problems \cite{12labella2006division}, we adopt an efficiency measure which is the ratio of the total income derived during solving the problem to the costs incurred during this period. In our particular variant of the foraging problem, we define income as the net prey content removed during the $\mathbf{N}$ time step run of the simulation. Cost incurred is the total energy spent by the robots in broadcasting invite signals. Assuming that the power used up while transmitting an invite during one time step, $\mathbf{P}$ is constant, the heuristic efficiency $\nu$ during one run of the algorithm is given as
     \begin{equation}
     \nu = \frac {\text{Total prey content}}{P \times \Sigma_{robots}\text{Invite duration}}
     \end{equation}
   \subsection{Simulation results}
    The simulation is run in the MATLAB simulation environment in a field containing 60 robots which are assigned to clear prey whose cumulative content is 6000 within a a given simulation time of  $\mathbf{N} = 1500$. The robots are assumed to take discrete steps of length, $\mathbf{l}=0.5$. The invite broadcast range is 30 and the power $\mathbf{P}$ used in transmitting an invite during one time step is 0.05. Prey are assumed to be small circular patches of radius 1. There are  two types of prey, small ones which contain 50 quantity of prey and large ones which contain 2900. The prey are randomly dispersed within the search space bounded by an imaginary edge length of 40. The robots are placed at two distinct patterns across the field as shown in Fig.\ref{config1} and Fig.\ref{config2}. Each configuration is simulated for 500 runs and two cumulative performance measures, Fig.\ref{perf1} and Fig.\ref{perf2}, are reported.
   \subsection{Discussion}

   \subsection*{Case A: Robots dispersed from two locations into field}
   As shown in Fig.\ref{config1}, the robots are released from two opposite corners into the field at the start of the simulation. The percentage prey grazed is measured for all three heuristics over five hundred runs. We observe from Fig.\ref{perf1config1} that the Hunger-loneliness heuristic removes fifty percent of the total prey content for over four hundred runs. The Random search-immediate invite heuristic does equivalently as well for only two hundred and fifty runs. However, as shown in Fig.\ref{perf2config1}, it is the performance measure $\nu$  that shows that Random search-immediate invite heuristic tends to make the apparent gains in prey removal an expensive affair due to the energy expended in blindly making invites.

   \subsection*{Case B: Robots dispersed from four locations into field }
\begin{figure*}
   \centering
   \subfigure [Case A: Robots dispersed from two locations into the field.]{
   \includegraphics[scale=.5]{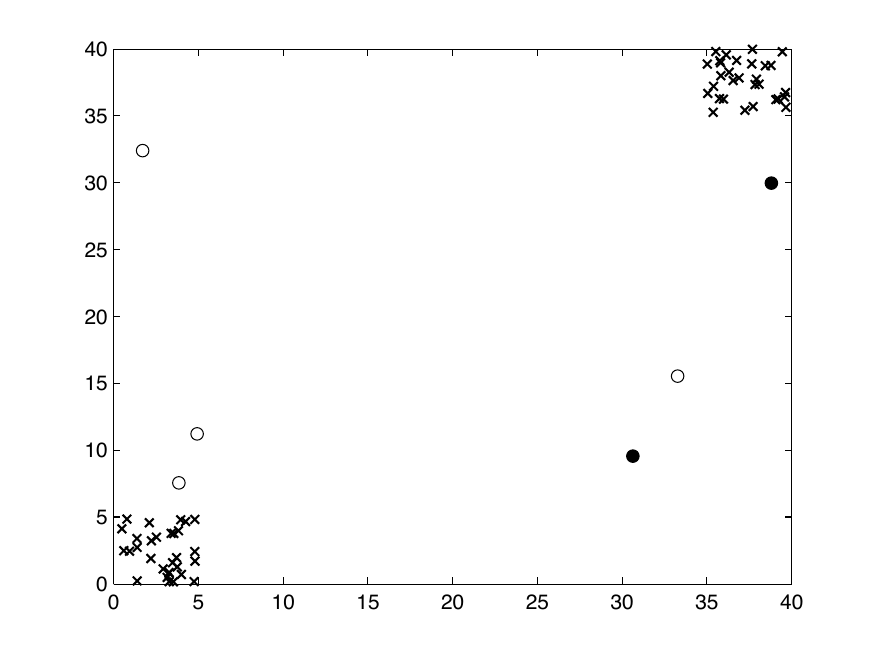}
    \label{config1}
   }
   \subfigure [Case B: Robots dispersed from four locations into the field.]{
   \includegraphics[scale=.5]{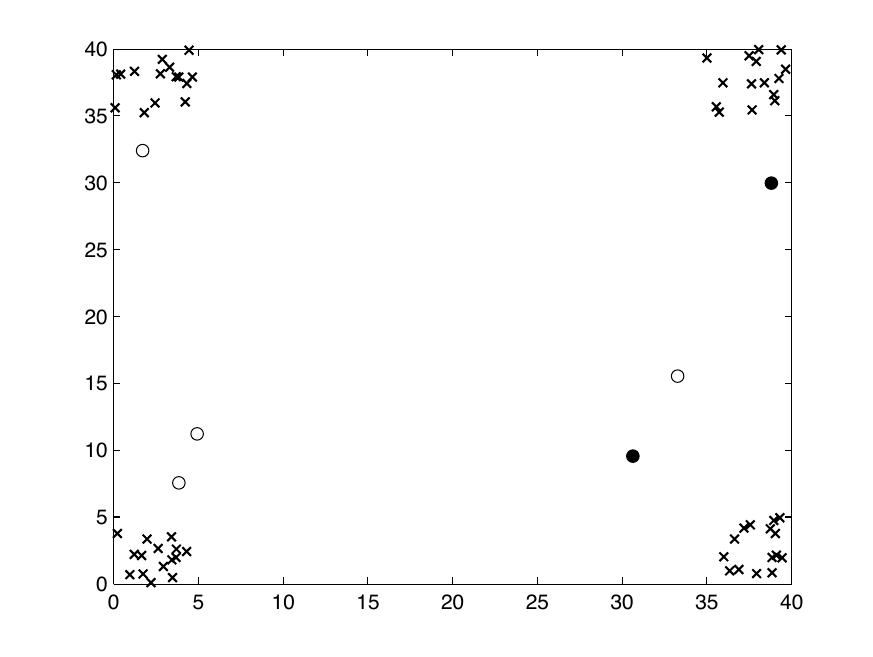}
     \label{config2}
   }
   \caption{Various robots configurations.}
   \label{robotConfig}
   \end{figure*}

\begin{figure*}
   \centering
   \subfigure [Case A]{
   \includegraphics[scale=.5]{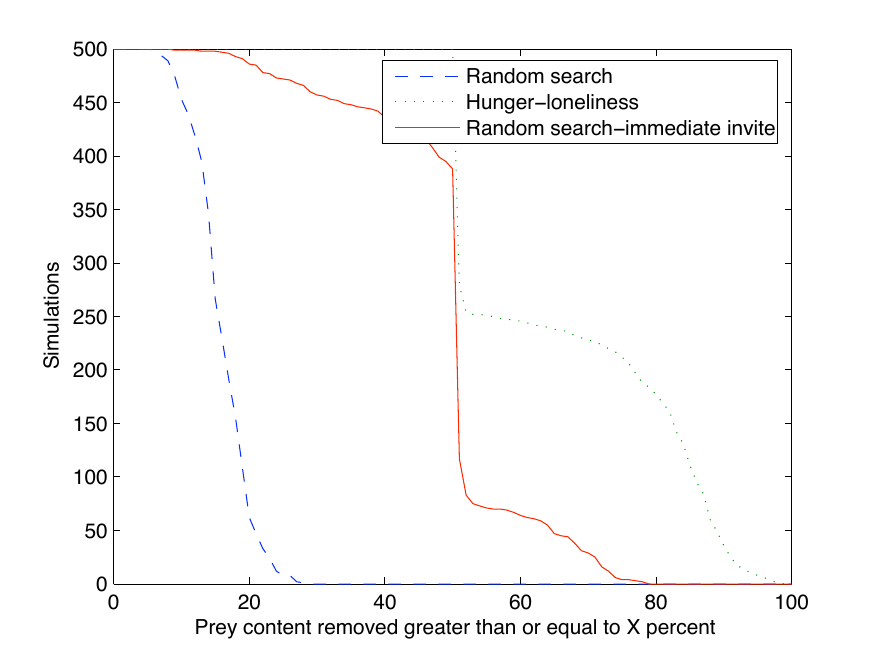}
     \label {perf1config1}
   }
   \subfigure [Case B]{
   \includegraphics[scale=.5]{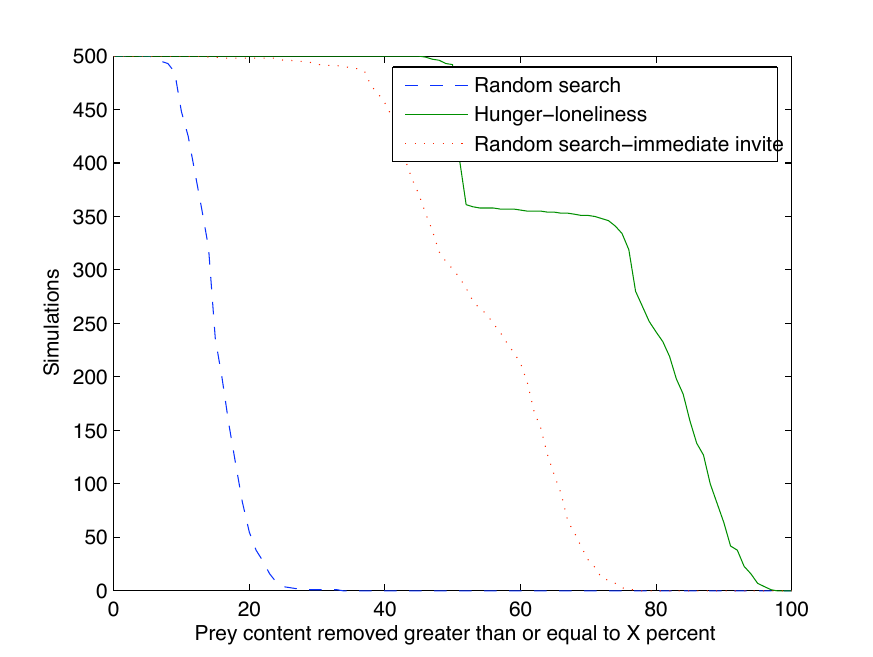}
     \label{perf1config2}
   }
  \caption{Total prey content removed for five hundred runs of all three heuristics.}
   \label{perf1}
   \end{figure*}

\begin{figure*}
   \centering
   \subfigure [Case A]{
   \includegraphics[scale=.5]{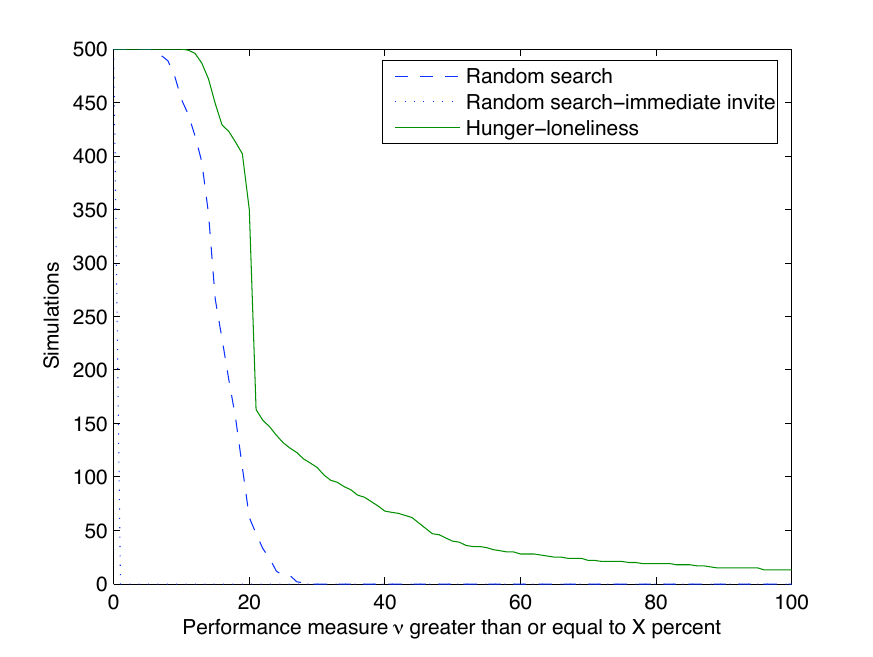}
   \label {perf2config1}
   }
   \subfigure [Case B]{
   \includegraphics[scale=.5]{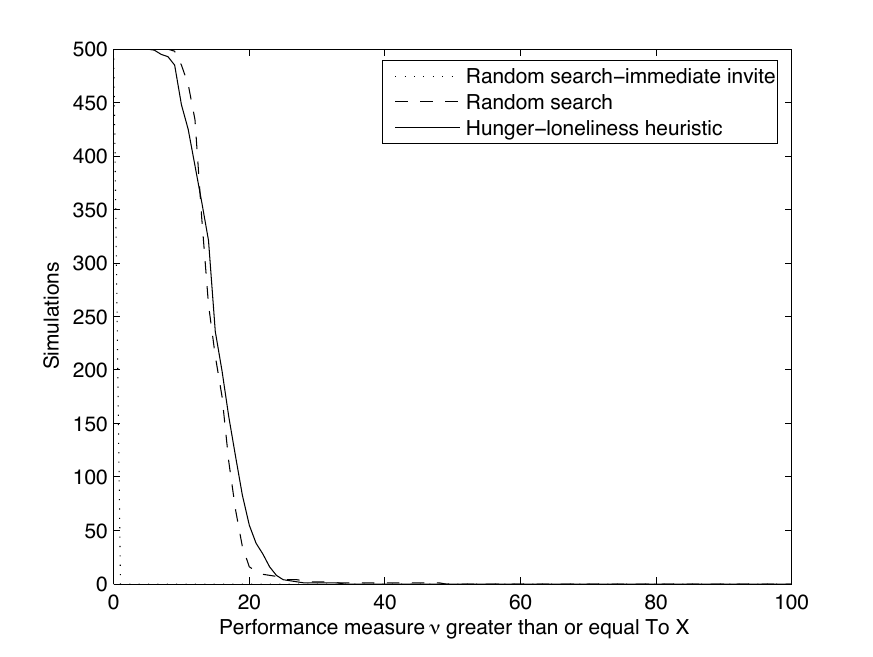}
   \label {perf2config2}
   }
   \caption{The values $\nu$ for five hundred runs of all three heuristics.}

   \label{perf2}
   \end{figure*}

  Fig.\ref{config2} shows the robots split up into groups of fifteen. Each group is released into the field from one of the four corners. The empirical analysis of performance is repeated. Although, the percentage of prey removed (Fig.\ref{perf1config2}) does not show any appreciable change between random search-immediate invite heuristic and hunger-loneliness heuristic, Fig. \ref{perf2config2} shows the strong improvement in hunger-loneliness heuristic when it comes to balancing the needs of conserving energy and removing prey.

     \section{CONCLUSION}
     We presented a new metaheuristic that adopts inspiration from emotions observed in social animals to enable a swarm of robots self-organize
     to exhibit emergent behavior in the form of removal of prey scattered randomly in the search space. There are several limitations that merit further exploration. These include utilizing a more detailed robot model where collision avoidance and robot dynamics are taken into account. In future work, we would also like to explore if it is possible to integrate a form of learning where the values of Hunger ($\mathbf{H}$) and Loneliness ($\mathbf{L}$) parameters at which the emotional state switch takes place can depend on the configuration of the environment.

\end{document}